%% file: paper.tex
\documentclass[sigconf]{acmart}
\def\BibTeX{{\rm B\kern-.05em{\sc i\kern-.025em b}\kern-.08emT\kern-.1667em\lower.7ex\hbox{E}\kern-.125emX}}

\usepackage{amsmath,amssymb}
\usepackage{graphicx,colortbl}
\usepackage{subfigure}
\usepackage{multirow}
\usepackage[boxed]{algorithm2e}
\usepackage{ctable}
\usepackage{hhline}
\usepackage{xcolor}
\usepackage{colortbl}

\begin{document}

\title{Understanding Behavior of Clinical Models under Domain Shifts} %Can Deep Clinical Models Handle Real-World Label Distribution Shifts?

% The "author" command and its associated commands are used to define the authors and their affiliations.
% Of note is the shared affiliation of the first two authors, and the "authornote" and "authornotemark" commands
% used to denote shared contribution to the research.
\author{Jayaraman J. Thiagarajan}
\affiliation{%
  \institution{Lawrence Livermore National Labs}
}
 \email{jjayaram@llnl.gov}

\author{Deepta Rajan}
\affiliation{%
  \institution{IBM Research Almaden}}
\email{drajan@us.ibm.com}

\author{Prasanna Sattigeri}
\affiliation{%
	\institution{MIT-IBM Watson AI Lab}}
\email{psattig@us.ibm.com}

\renewcommand{\shortauthors}{J. J. Thiagarajan, et al.}

%
% The abstract is a short summary of the work to be presented in the article.
\begin{abstract}
%We will make the world better!
The hypothesis that computational models can be reliable enough to be adopted in prognosis and patient care is revolutionizing healthcare. Deep learning, in particular, has been a game changer in building predictive models, thus leading to community-wide data curation efforts. However, due to inherent variabilities in population characteristics and biological systems, these models are often biased to the training datasets. This can be limiting when models are deployed in new environments, when there are systematic domain shifts not known a priori. In this paper, we propose to emulate a large class of domain shifts, that can occur in clinical settings, with a given dataset, and argue that evaluating the behavior of predictive models in light of those shifts is an effective way to quantify their reliability. More specifically, we develop an approach for building realistic scenarios, based on analysis of \textit{disease landscapes} in multi-label classification. Using the openly available MIMIC-III EHR dataset for phenotyping, for the first time, our work sheds light into data regimes where deep clinical models can fail to generalize. This work emphasizes the need for novel validation mechanisms driven by real-world domain shifts in AI for healthcare.
\end{abstract}




%
%\begin{teaserfigure}
%	\includegraphics[width=\textwidth]{sampleteaser}
%	\caption{Seattle Mariners at Spring Training, 2010.}
%	\Description{Enjoying the baseball game from the third-base seats. Ichiro Suzuki preparing to bat.}
%	\label{fig:teaser}
%\end{teaserfigure}

%
% This command processes the author and affiliation and title information and builds
% the first part of the formatted document.
\maketitle

\section{Introduction}
\label{sec:intro}
\input{intro.tex}
\section{Dataset Description}
\input{data.tex}

\section{Analysis of Deep Clinical Models}
\input{model.tex}
\section{Characterizing Domain Shifts}
\label{sec:scenario}
\input{scenario.tex}

\section{Empirical Results}
\input{results.tex}
\bibliographystyle{ACM-Reference-Format}
\bibliography{paper.bib}

\end{document}

%% file: intro.tex
The role of automation in healthcare and medicine is steered by both the ever-growing need to leverage knowledge from large-scale, heterogeneous information systems, and the hypothesis that computational models can actually be reliable enough to be adopted in diagnostics. Deep learning, in particular, has been a game changer in this context, given recent successes with both electronic health records (EHR) and raw measurements (e.g. ECG, EEG) ~\cite{rajkomar2018scalable}, ~\cite{rajan2018generative}. 
While designing computational models that can account for complex biological phenomena and all interactions during disease evolution is not possible yet, building predictive models is a significant step towards improving patient care. There is a fundamental trade-off in predictive modeling with healthcare data. While the complexity of disease conditions naturally demands expanding the number of predictor variables, this often makes it extremely challenging to identify reliable patterns in data, thus rendering the models heavily biased. In other words, despite the availability of large-sized datasets, we are still operating in the small data regime, wherein the curated data is not fully representative of what the model might encounter when deployed.

In this paper, we consider a variety of population biases, label distribution shifts and measurement discrepancies that can occur in clinical settings, and argue that evaluating the behavior of predictive models in light of those shifts is crucial to understanding strengths and weaknesses of predictive models. To this end, we consider the problem of phenotyping using EHR data, characterize different forms of discrepancies that can occur between train and test environments, and study how the prediction capability varies across these discrepancies. Our study is carried out using the openly available MIMIC-III EHR dataset \cite{johnson2016mimic} and the state-of-the-art deep ResNet model~\cite{BaiTCN2018}. The proposed analysis provides interesting insights about which discrepancies are challenging to handle, which in turn can provide guidelines while deploying data-driven models.

%% file: data.tex
All experiments are carried out using the MIMIC-III dataset \cite{johnson2016mimic}, the largest publicly available database of de-identified EHR from ICU patients. It includes a variety of data types such as diagnostic codes, survival rates, length of stay etc., yielding a total of $76$ measurements. For our study, we focus on the task of acute care phenotyping that involves retrospectively predicting the likely disease conditions for each patient given their ICU measurements. Preparation of such large heterogeneous records involve organizing each patient's data into episodes containing both time-series events as well as episode-level outcomes like diagnosis, mortality etc.
%Here, the time-series data includes a total of 17 measurements such as capillary refill rate, systolic, diastolic and mean blood pressure, glucose, heart rate, oxygen saturation, temperature, height, weight, pH, and Glascow Coma Scale (GCS) parameters like eye opening, motor and verbal response. 
Phenotyping is typically a multi-label classification problem, where each patient can be associated with several disease conditions.  The dataset contains $25$ disease categories, $12$ of which are critical such as respiratory/renal failure, $8$ are chronic conditions such as diabetes, atherosclerosis, with the remaining $5$ being 'mixed' conditions such as liver infections. Note that, in our study, we reformulate the phenotyping problem as a binary classification task of detecting the presence or absence of a selected subset of diseases.

\begin{figure*}[t]
	\centering
	\centerline{\includegraphics[width=0.99\linewidth]{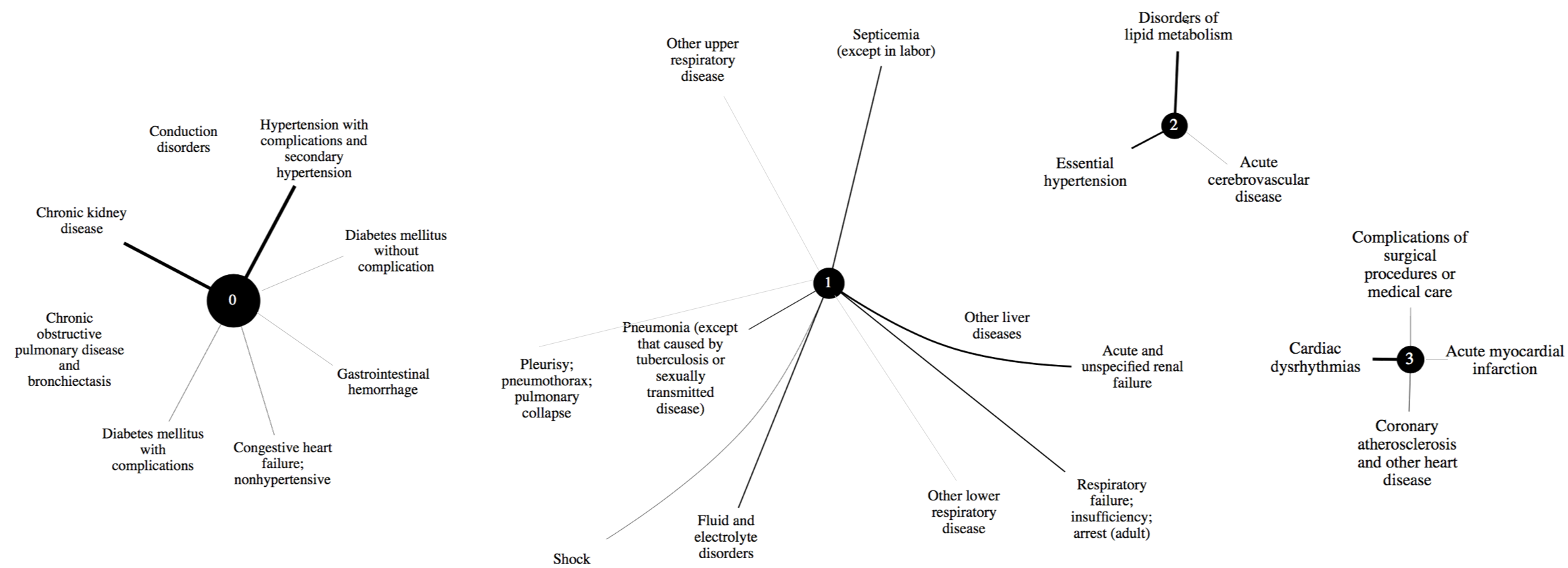}}
	\caption{Constructing \textit{disease landscapes}: Using information decomposition on the MIMIC-III EHR dataset reveals the grouping of disease outcomes.}
	\label{fig:disease_correl}
	\vspace{-10pt}
\end{figure*}

%Describe the dataset and task. this can be fairly detailed.
%To control errors in the phenotyping labeling, only hospital admissions with a single ICU stay were considered, and the labels were curated by mapping a subset of ICD-9 codes to the ICU stays using hospital visit identifiers. 
%As part of data pre-processing, the time-series measurements were further transformed into a $76$-dimensional vector at each time-step, through expansion of categorical measurements into multiple binary variables. 
% This design is motivated by the need to understand variabilities in the population that are harder to generalize, thus providing a more realistic evaluation of clinical models. 
%The sizes of the datasets from all our scenarios are shown in . In the next section, we will describe the proposed approach for characterizing domain shifts through scenario design.
%providing patient summaries, compiling specific cohorts, finding similarities between patients, and performing risk adjustment
%\cellcolor{gray!20}
%Each scenario is comprised of a pair of source and target domains, represented by partitions of the ICU patient episodes, that are chosen based on a scenario-specific constraint (e.g. gender distribution).
%- AUROC to detect Heart failure using 12-month window auroc of 0.78, and using 18-month window = 0.88.

%% file: model.tex
%%This section discusses challenges in building data-driven solutions for healthcare, current art in deep learning based clinical modeling, mainly using EHR data, as well as an overview of  evaluation strategies used in practice.
%Modeling such data is constrained by inherent data-related challenges such as long range temporal dependencies, missing measurements, and varying sampling rates.
Clinical time-series modeling broadly focuses on problems such as anomaly detection, tracking progression of specific diseases or detecting groups of diseases. Deep learning provides a data-driven approach to exploiting relationships between a large number of input measurements while producing robust diagnostic predictions. In particular, Recurrent Neural Networks (RNN) was one of the earlier solutions for dealing with sequential data in tasks such as discovering phenotypes and detecting cardiac diseases \cite{rajan2018generative}, etc. The work by~\cite{rajkomar2018scalable} extensively benchmarked the effectiveness of deep learning techniques in analyzing EHR data. More recently, convolutional networks \cite{BaiTCN2018}, and attention models \cite{song2017attend} have become highly effective alternatives to sequential models. In particular, deep Residual Networks with $1D$ convolutions (ResNet-$1D$), have produced state-of-the-art performance in many clinical tasks~\cite{BaiTCN2018}. Hence, in this work, all studies are carried out using this architecture.

\noindent \textbf{Architecture}: We use a ResNet-$1D$ model with $7$ residual blocks which transform the raw time-series input into feature representations. Each residual block is made up of two 1-D convolution and batch normalization layers with a kernel size of $3$ and $64$ or $256$ filters, a dropout and a leaky-ReLU activation layer. Further, a $1-D$ max pooling layer is used after the third and seventh residual blocks to aggregate the features. Additional parameters include: an Adam optimizer with a learning rate of $0.001$, and a batch size of $128$ with an equal balance of positive and negative samples. 

\noindent \textbf{Proposed Work}: An important challenge with the design of deep clinical models is the lack of a holistic understanding of their behavior under the wide-range of discrepancies that can occur between train and test environments. In particular, we consider the following domain shifts -- (i) \textit{population biases} such as age, gender, race etc.; (ii) \textit{label distribution shifts} such as novel disease conditions at test time, presence of combinations of observed diseases etc.; and (iii) \textit{measurement discrepancies} such as noisy labels, missing measurements, variations in sampling rate etc. Understanding how much these discrepancies impact a model's generalization performance is critical to qualitatively understanding its strengths and expected regimes of failure, and thus provide guidelines for effective deployment. Conceptually, this amounts to quantifying the prediction uncertainties arising due to the lack of representative data and inherent randomness from clinical data.

%% file: scenario.tex
In this section, we describe our approach for emulating a variety of domain shifts that commonly occur in a clinical setting, which will be then used for evaluating the deep 1D-ResNet model in the next section. For each of the cases, we construct a source and a target dataset, where the data samples are chosen based on a discrepancy-specific constraint (e.g. gender distribution, males used in source and females in target). The positive class corresponds to the presence of a selected subset of diseases (e.g. organ specific conditions), while the negative class indicates their absence in the patients. Note, we ensured that the resulting datasets are balanced, and had positive-negative split of $\sim 60\%-40\%$. Before we describe the different discrepancies considered in our study, we will briefly discuss the information decomposition technique~\cite{steeg2016information} that will be used to explore the landscape of diseases in order to pick a subset of related conditions for defining the positive class.
%The list of domain shifts considered and the dataset sizes are shown in Table \ref{table:dataset}. 

\noindent \textbf{Exploring Disease Landscapes}: We utilize \textit{information sieve} \cite{steeg2016information}, a recent information-theoretic technique for identifying latent factors in data that maximally describe the \textit{total correlation} between variables. Denoting a set of multivariate random variables by $\tilde{X} = \{\tilde{X}_i\}_{i=1}^d$, dependencies between the variables can be quantified using the multivariate mutual information, also known as total correlation (TC), which is defined as
\begin{equation}
TC(\tilde{X}) = \sum_{i=1}^d H(\tilde{X}_i) - H(\tilde{X}),
\end{equation}
where $H(\tilde{X}_i)$ denotes the marginal entropy. This quantity is non-negative and zero only when all the $\tilde{X}_i$'s are independent. Further, denoting the latent source of dependence in $\tilde{X}$ by $\tilde{Z}$, we can define the conditional $TC(\tilde{X}|\tilde{Z})$, i.e., the residual total correlation after observing $\tilde{Z}$. Hence, we solve the problem of searching for $\tilde{Z}$ that minimizes
$TC(\tilde{X}|\tilde{Z})$. Equivalently, we can define the reduction in $TC$ after conditioning on $\tilde{Z}$ as $TC(\tilde{X};\tilde{Z}) = TC(\tilde{X})-TC(\tilde{X}|\tilde{Z})$. The optimization begins with $\tilde{X}$, constructs $\tilde{Z}_0$ to maximize $TC(\tilde{X};
\tilde{Z}_0)$. Subsequently, it computes the remainder information not explained by
$\tilde{Z}_0$, and learns another factor, $\tilde{Z}_1$, that infers additional dependence
structure. This procedure is repeated $k$ times until the entirety of
multivariate mutual information in $\tilde{X}$ is explained.

Denoting a labeled dataset using the tuple $(\mathbf{X}, \mathbf{Y})$, where $\mathbf{X} \in \mathbb{R}^{N \times D}$ is the input data with $D$ variables, and $\mathbf{Y} \in \mathbb{R}^{N \times d}$ is the label matrix with $d$ different categories, we use \textit{information sieve} to analyze the outcome variables. In other words, we identify latent factors to describe the total correlation in $\mathbf{Y}$ and the resulting hierarchical decomposition is referred as the \textit{disease landscape}. The force-based layout in Figure \ref{fig:disease_correl} provides a holistic view of the landscape for the entire MIMIC-III dataset. Here, each circle corresponds to a latent factor and their sizes indicate the amount of total correlation explained by that factor, and the thickness of edges reveal contributions of each outcome variable to that factor. 

\subsection{Population Biases}
Clinical datasets can be imbalanced as to population characteristics. 

\noindent \textbf{(i) Age}:  (a) \textit{Older-to-Younger}: Source comprises of patients who are $60$ years and above, and the target includes patients who are below $60$ years of age; (b) \textit{Younger-to-Older}: This represents the scenario with source containing patients younger than $60$. Once the population is divided based on the chosen age criteria, we pick a cluster of diseases in the source's disease landscape as the positive class. However, due to the domain shift, the landscape can change significantly in the target. Hence, in the target, we recompute the landscape and consider the source conditions as well as diseases strongly correlated to any of those conditions as positive.

\noindent \textbf{(ii) Gender}: Similar to the previous case, we emulate two cases,  \textit{Male-to-Female}, i.e. source data consists only of patients who identify as male and target has only patients who are female. While, the second scenario  \textit{Female-to-Male} is comprised of female-only source data and male-only target data.

\noindent \textbf{(iii) Race}: We emulate \textit{White-to-Minority}, that contains prevalent racial groups such as white American, Russian, and European in the source, while constructing the target domain comprising of patients belonging to minority racial groups like Hispanic, South America, African, Asian, Portuguese, and others marked as unknown.

\subsection{Label Distribution Shifts}

When models are trained to detect the presence of certain diseases, they can be ineffective when novel disease conditions or variants of existing conditions, previously unseen by the model, appear. 

\noindent \textbf{(i) Novel Diseases at Test Time}: (a) \textit{Resp-to-CardiacRenal}: In the source, we detect the presence or absence of diseases from cluster $1$ (Figure \ref{fig:disease_correl}), while the target requires detection of the presence of at least one of the conditions from cluster $1$ or unobserved conditions in cluster $0$; (b) \textit{Cerebro-to-CardiacRenal}: This is formulated as detecting diseases from cluster $2$ in the source dataset, while expanding the disease set in the target by including cluster $0$. 

\noindent \textbf{(ii) Dual-to-Single}: As a common practice, lab tests for two diseases could be conducted together based on the likelihood of them co-occurring, giving rise to a dataset containing patients with both diseases. However, one would expect to detect the presence of even one of those diseases. Here, source includes patients associated with two disease conditions simultaneously, while target includes patients who present only one of the two. Note, we do not construct the landscapes for this case, instead, we divide patients into four groups, namely: those that have only cardiac diseases, only renal diseases, neither cardiac or renal diseases, and, both cardiac and renal diseases, and build the source-target pair.

\noindent \textbf{(iii) Single-to-Dual}: Similar to the previous case, an alternate situation could occur with models having to adapt to predicting patients that have two diseases, while having been trained on patients who were diagnosed with only one of them. Such a scenario is explored using a source dataset comprising of patients diagnosed as having either cardiac only or respiratory only disease, and a target dataset with patients that have both diseases.

\subsection{Measurement Discrepancies}

\noindent \textbf{(i) Noisy Labels}: Variabilities in diagnoses between different experts is common in clinical settings \cite{valizadegan2013learning}. Consequently, it is important to understand the impact of those variabilities on behavior of the model, when adopted to a new environment. We extend the \textit{Resp-to-CardiacRenal} case by adding uncertainties to the diagnostic labels in the source data and study its impact on the target. In particular, we emulate two such scenarios by randomly flipping $10$\% and $20$\% of the labels respectively. 

\noindent \textbf{(ii) Sampling Rate Change}: Another typical issue with time-series data is the variability in sampling rates. We create a source-target pair based on the \textit{Resp-to-CardiacRenal} scenario with measurements collected at different sampling rates. The source uses samples at every $96$ hours while the target is sampled at $48$ hours. 

\noindent \textbf{(iii) Missing Measurements}: Clinical time-series is plagued by missing measurements during deployment, often due to resource or time limitations. We emulate this discrepancy based on the \textit{Dual-to-Single} scenario, wherein we assume that the set of measurements, \textit{pH}, \textit{Temperature}, \textit{Height}, \textit{Weight}, and all \textit{Verbal Response GCS} parameters, were not available in the target dataset. We impute the missing measurements with zero and this essentially reduces the dimensionality of valid measurements from $76$ to $41$ in the target.

%% file: results.tex
\begin{figure*}[t]
	\centering
	\subfigure[Population Biases]{
		\includegraphics[height=.18\linewidth]{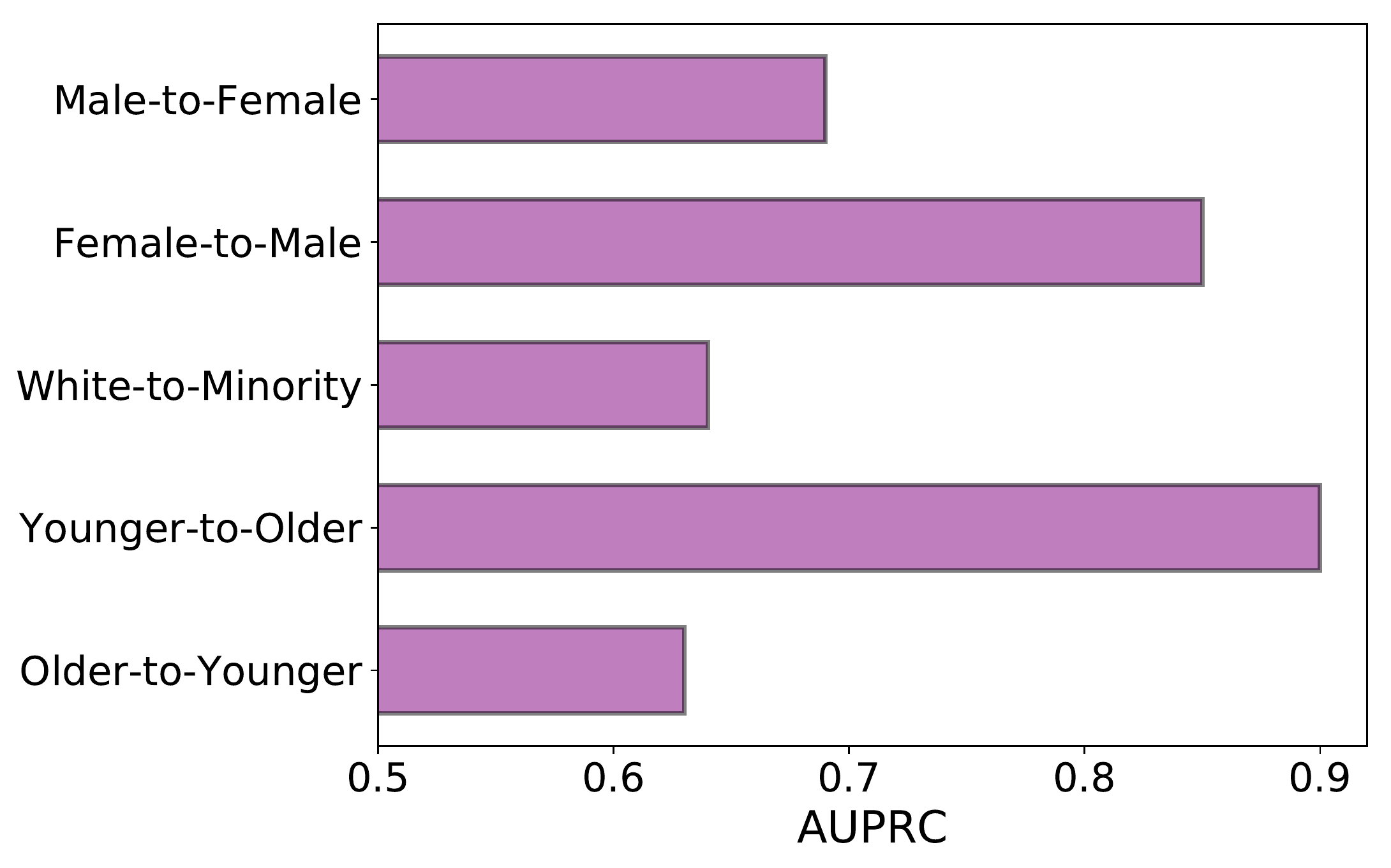} \label{fig:pop}
	}
	\subfigure[Label Distribution Shifts]{
		\includegraphics[height=.18\linewidth]{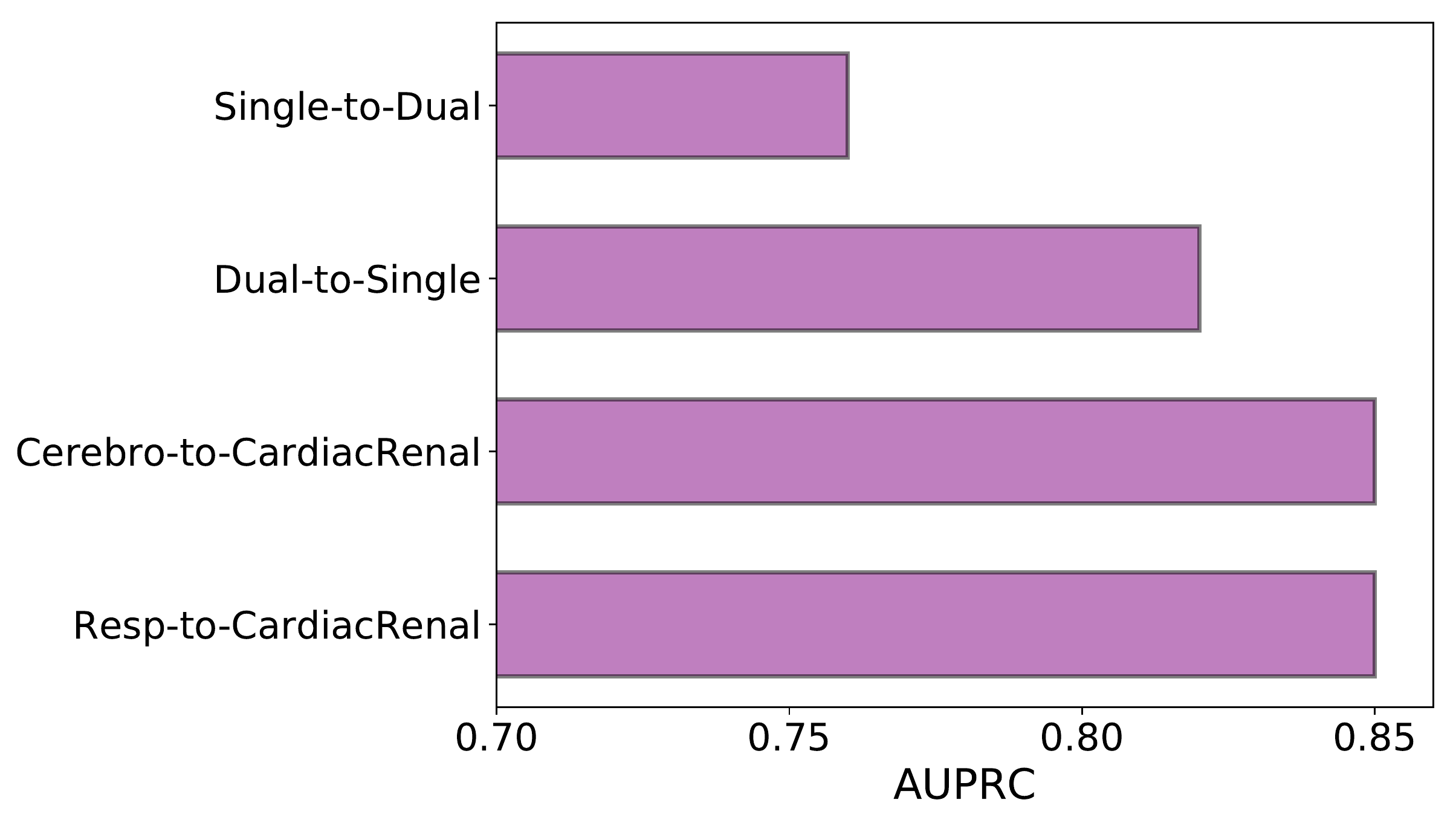} \label{fig:lab}
	}
	\subfigure[Measurement Discrepancies]{
		\includegraphics[height=.18\linewidth]{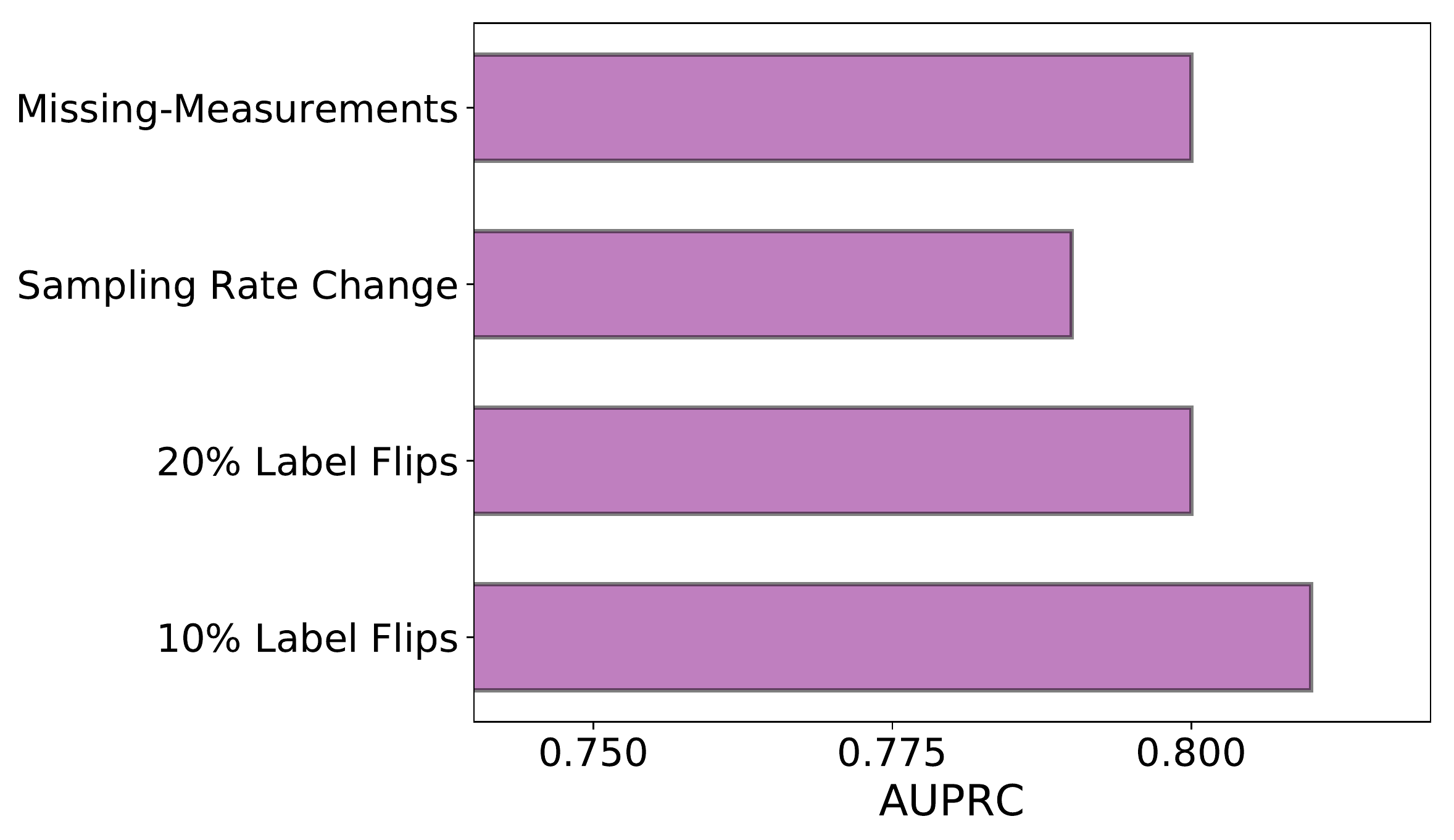} \label{fig:meas}
	}
	\caption{Results from our empirical study on generalizability of the ResNet-$1D$ model with complex domain-shifts.}
	\label{fig:results}
\end{figure*}

\small
\begin{table*}[t]
	\caption{Disease conditions considered in source-target pairs for scenarios where the model demonstrated poor generalization.}
	\label{tab:disease}
	\renewcommand*{\arraystretch}{1.3}
	\begin{tabular}{c|p{6cm}|p{6cm}}
		\hline
		\textbf{Shift}                     & \textbf{Source Diseases}                                                                                                   & \textbf{Target Diseases}                                                                                                                                                     \\
		\hline
		Older-to-Younger                   & Coronary atherosclerosis, Disorders of lipid metabolism, Essential hypertension & Coronary atherosclerosis, Disorders of lipid metabolism, Essential hypertension, Chronic kidney disease, Secondary hypertension \\ \hline
		Male-to-Female                     & Dysrhythmia, Congestive heart failure             & Dysrhythmia, Congestive Heart Failure, Coronary atherosclerosis      \\ \hline
		White-to-Minority                  & Dysrhythmia, Conduction disorder, Congestive heart failure & Dysrhythmia, Conduction disorder, Congestive heart failure, Chronic kidney disease, Secondary hypertension, Diabetes \\ \hline
		Single-to-Dual & Cardiac-only or Renal-only& Both Cardiac and Renal Diseases \\
		\hline                                                                                                                      
	\end{tabular}
\end{table*}
\normalsize

%While AUPRC is a common choice in clinical model evaluation, MCC can be a more balanced measure for binary classification problems, even when the classes are of different sizes. The Matthews Correlation Coefficient is computed as follows:
%$$
%MCC = \frac{tp * tn - fp * fn}{\sqrt{(tp + fp)(tp + fn)(tn + fp)(tn + fn)}},
%$$ where $tp$, $tn$, $fp$, $fn$ denote the total number of true positives, true negatives, false positives and false negatives.
In this section, we present empirical results from our study -- performances of the ResNet-$1D$ model trained on the source data when subjected to perform under a variety of discrepancies. Following common practice, we considered the weighted average AUPRC metric. Figure \ref{fig:results} illustrates the performance of the ResNet-$1D$ model for the three categories of discrepancies. We report the weighted AUPRC score for each of the cases by training the model on source and testing on the target. Broadly, these results characterize the amount of uncertainty for even a sophisticated ML model, when commonly occurring discrepancies are present. 

The first striking observation is that population bias leads to significant performance variability, wherein the AUPRC score varies in a wide range between $0.65$ and $0.9$. Note that, by identifying disparate sets of disease conditions, the reported results correspond to the \textit{worst-case performance} for each scenario. In particular, we observe that the racial bias \textit{White-to-Minority}, the gender bias \textit{Male-to-Female}, and the age bias \textit{Older-to-Younger} demonstrate challenges in generalization. This sheds light on the fact that manifestation of different disease conditions in the younger population shows higher variability compared to older patient groups, and hence a model overfit to the older case does not generalize to the former. Table~\ref{tab:disease} lists the set of source and target diseases for the cases with poor generalization. In the \textit{Older-to-Younger} scenario, it is observed that the disease landscape of older patients reveals strong co-occurrence between \textit{Coronary atherosclerosis}, \textit{Disorders of lipid metabolism}, and \textit{Essential hypertension}. In contrast, when one of these diseases occur in younger patients, it is often accompanied by other conditions such as \textit{Secondary hypertension} and \textit{Chronic kidney disease}. This systematic shift challenges pre-trained models to generalize well to the target. Similarly, in the case of \textit{White-to-Minority}, while \textit{Congestive Heart Failure} commonly manifests with \textit{Dysrhythmia} in a predominantly white population, additional conditions such as \textit{Secondary Hypertension} and \textit{Diabetes} co-occur among minorities.

In the case of label distribution shifts, when there is no population bias, surprisingly the model is able to generalize well even when unseen disease conditions are present in the target. This is evident from the high AUPRC scores for both \textit{Resp-to-CardiacRenal} and \textit{Cerebro-to-CardiacRenal}. However, the EHR signatures for patients that present subsets or supersets of diseases observed in source are challenging to handle. In particular, detecting the presence of both cardiac and renal conditions using source data with patients who had only either of the diseases as in \textit{Single-to-Dual}. This clearly shows the inherent uncertainties in biological systems (often referred to as \textit{aleatoric} uncertainties) that cannot be arbitrarily reduced by building sophisticated ML models.

Finally, with respect to measurement discrepancies, it is widely believed that quality of labels in the source data is highly critical. However, surprisingly, we observe that with limited noise in the labels (\textit{10\% Label Flips}), the performance degradation is minimal. However, as the amount of noise increases (\textit{20\% Label Flips}), there is further degradation. Another important observation is that sampling rate change has a negative effect on the generalization performance and simple imputation does not suffice (we adopt a strategy where value from the last time-step is repeated). In comparison, the model is fairly robust to missing measurements.